%% file: main.tex
\documentclass{article}

\usepackage{iclr2025_conference,times}

\input{preamble/packages}
\input{preamble/macros}
\hypersetup{
	colorlinks=false,
	citebordercolor={0 1 0},
	linkbordercolor={1 0 0},
	urlbordercolor={1 0 0}
}

\title{ConsistNav: Closing the Action Consistency Gap in Zero-Shot Object Navigation with Semantic Executive Control}

\author{%
\normalfont\footnotesize
\setlength{\tabcolsep}{0pt}
\renewcommand{\arraystretch}{1.16}
\begin{tabular*}{0.94\textwidth}{@{\extracolsep{\fill}}>{\raggedright\arraybackslash}p{0.28\textwidth}>{\raggedright\arraybackslash}p{0.28\textwidth}>{\raggedright\arraybackslash}p{0.28\textwidth}}
\\[0.1em]
\textbf{Haosen Wang}\textsuperscript{*} &
\textbf{Zhenyang Li}\textsuperscript{*} &
\textbf{Yinqiang Zhang} \\
Sun Yat-sen University &
The University of Hong Kong &
The University of Hong Kong \\
\\[0.5em]
\textbf{Zongqi He} &
\textbf{Lutao Jiang} &
\textbf{Kai Li} \\
The University of Hong Kong &
Hong Kong University of Science and Technology (Guangzhou) &
City University of Hong Kong \\
& & \\[0.5em]
\textbf{Yizhou Zhao} &
\textbf{Liaoyuan Fan} &
\textbf{WenjianHou} \\
Carnegie Mellon University &
The University of Hong Kong &
Sun Yat-sen University \\
& & \\[0.5em]
\textbf{Tingbang Liang} &
\textbf{Yibin Wen} &
\textbf{Defeng Gu}\textsuperscript{\textdagger} \\
Sun Yat-sen University &
Sun Yat-sen University &
Sun Yat-sen University \\
& &
\end{tabular*}
\\[0.45em]
\textsuperscript{*}Equal contribution.
\quad
\textsuperscript{\textdagger}Corresponding author.
}

\iclrfinalcopy

\begin{document}

\maketitle

\input{sections/abstract}
\input{sections/introduction}
\input{sections/related_work}
\input{sections/method}
\input{sections/experiment}
\input{sections/conclusion}

\bibliographystyle{iclr2025_conference}
\bibliography{references}

\end{document}

%% file: preamble/packages.tex
\usepackage[utf8]{inputenc}
\usepackage[T1]{fontenc}
\usepackage{microtype}
\usepackage{xcolor}
\usepackage{graphicx}
\usepackage{amsmath}
\usepackage{amssymb}
\usepackage{amsfonts}
\usepackage{booktabs}
\usepackage{array}
\usepackage{nicefrac}
\usepackage{url}
\usepackage{hyperref}
\let\ConsistNavOriginalRef\ref
\renewcommand{\ref}[1]{%
  \begingroup
  \hypersetup{pdfborder={0 0 0}}%
  \hyperref[#1]{%
    {\setlength{\fboxsep}{0pt}\setlength{\fboxrule}{0.4pt}%
     \fcolorbox{red}{white}{\normalcolor\ConsistNavOriginalRef*{#1}}}%
  }%
  \endgroup
}
\usepackage{enumitem}
\usepackage{multirow}
\usepackage{pifont}
\usepackage[ruled,vlined,linesnumbered]{algorithm2e}
\usepackage{wrapfig}
\usepackage{adjustbox}
\usepackage{placeins}

%% file: preamble/macros.tex
\newcommand{\stSearch}{\textsc{Search}}
\newcommand{\stSuspect}{\textsc{Suspect}}
\newcommand{\stApproach}{\textsc{Approach}}
\newcommand{\stVerify}{\textsc{Verify}}
\newcommand{\stFinalApproach}{\textsc{Final-Approach}}
\newcommand{\stFailover}{\textsc{Failover}}
\newcommand{\stSuccess}{\textsc{Success}}

\newcommand{\ours}{\textbf{ConsistNav}}

\newcommand{\SR}{SR}
\newcommand{\SPL}{SPL}

\newcommand{\tightcitealp}[1]{\mbox{\raisebox{0pt}[1.05ex][0.2ex]{\citealp{#1}}}}
\newcommand{\tightcitep}[1]{(\tightcitealp{#1})}

%% file: sections/abstract.tex
\begin{abstract}
Zero-shot object navigation has advanced rapidly with open-vocabulary detectors, image--text models, and language-guided exploration. However, even after current methods detect a plausible target hypothesis, the agent may still oscillate between exploration and pursuit, or abandon the object near success. We identify this failure mode as an action consistency gap: semantic evidence is repeatedly reinterpreted at each step without persistent commitment across the episode. We introduce ConsistNav, a training-free zero-shot ObjectNav framework built around a semantic executive composed of three coordinated modules: Finite-State Executive Controller stages target pursuit through guarded semantic phases; Persistent Candidate Memory accumulates cross-frame target evidence into stable object hypotheses; and Stability-Aware Action Control suppresses rotational stagnation, ineffective pursuit, and unverified stopping. This design changes neither the detector nor the low-level planner; instead, it controls when semantic evidence should influence navigation and when it should be suppressed or revisited. We conduct extensive experiments on HM3D and MP3D, where ConsistNav achieves state-of-the-art results among compared zero-shot ObjectNav methods and improves SR by 11.4\% and SPL by 7.9\% over the controlled baseline on MP3D. Ablation studies and real-world deployment experiments further demonstrate the effectiveness and robustness of the proposed executive mechanism.
\end{abstract}

%% file: sections/introduction.tex
\section{Introduction}
\label{sec:intro}

Object-goal navigation requires an embodied agent to find a target-category instance in an unseen environment and stop within the success radius~\tightcitep{batra2020objectnav}. In zero-shot ObjectNav (ZSON), open-vocabulary detectors, vision--language models, and language-guided frontier exploration have improved semantic search without task-specific policy training~(\tightcitealp{radford2021clip}; \tightcitealp{khandelwal2022clip}; \tightcitealp{majumdar2022zson}; \tightcitealp{yu2023l3mvn}; \tightcitealp{zhou2023esc}; \tightcitealp{yin2024sgnav}; \tightcitealp{yokoyama2024vlfm}; \tightcitealp{zhang2025apexnav}). However, many modular systems still optimize \emph{semantic evidence production} or \emph{frontier scoring} more directly than the conversion of evidence into sustained action~(\tightcitealp{yamauchi1997frontier}; \tightcitealp{zhou2023esc}; \tightcitealp{yin2024sgnav}; \tightcitealp{yokoyama2024vlfm}; \tightcitealp{zhang2025apexnav}). When semantic observations are repeatedly reinterpreted against newly scored frontiers, an agent may oscillate between exploration and pursuit, abandon a plausible target near success, or stop from an insufficiently verified viewpoint. We call this the \emph{action consistency gap}: useful evidence is observed, but the induced objective, commitment, and stop decision are not preserved through approach, verification, and termination.

We introduce ConsistNav, a training-free perception--planning--execution framework for ZSON centered on a semantic executive. The executive contains three coordinated modules. \emph{Finite-State Executive Controller} stages target pursuit through guarded semantic phases. \emph{Persistent Candidate Memory} accumulates cross-frame target evidence into stable object hypotheses. \emph{Stability-Aware Action Control} suppresses rotational stagnation, ineffective pursuit, and unverified stopping. Together, these modules convert open-vocabulary semantic evidence into temporally consistent action while leaving the detector, semantic mapper, and low-level planner unchanged.

We evaluate ConsistNav through controlled experiments under the same underlying stack, isolating the effect of executive control from changes in perception or planning capacity. Beyond standard Success Rate (SR) and Success weighted by Path Length (SPL), we analyze diagnostic failure categories that separate verified success from infeasible targets, unstable commitments, frontier exhaustion, timeouts, and missed targets. We further report real-world quantitative deployment results, showing that the same executive principles transfer beyond simulation.

Our contributions are threefold:
\begin{itemize}[leftmargin=1.5em]
    \item \textbf{Evidence-to-action diagnosis.} We formulate the \emph{action consistency gap} in ZSON, showing that retained semantic evidence can still produce unstable pursuit, weak verification, and premature stopping when commitment is not preserved through execution.
    \item \textbf{Executive semantic commitment.} We introduce a training-free semantic executive that combines persistent candidate memory, guarded finite-state control, and stability-aware action filtering to turn open-vocabulary evidence into stable, recoverable, and termination-aware navigation commitments.
    \item \textbf{Same-stack validation.} With the detector, mapper, frontier planner, and low-level planner fixed, ConsistNav achieves state-of-the-art results among compared ZSON methods, with ablations, failure diagnostics, and real-world deployment showing that the gains come from executive control rather than a stronger perception or planning backbone.
\end{itemize}

%% file: sections/related_work.tex
\section{Related Work}
\label{sec:related}

\subsection{Semantic Frontier Navigation}

Zero-shot ObjectNav increasingly relies on pretrained visual--language representations rather than category-specific navigation policies. CLIP-style goal embeddings and open-vocabulary detections enable navigation to unseen categories~(\tightcitealp{khandelwal2022clip}; \tightcitealp{majumdar2022zson}), while later systems add frontier maps, language priors, scene graphs, and multimodal commonsense reasoning~(\tightcitealp{gadre2023cows}; \tightcitealp{yokoyama2024vlfm}; \tightcitealp{yu2023l3mvn}; \tightcitealp{shah2022lmnav}; \tightcitealp{zhou2023esc}; \tightcitealp{yin2024sgnav}; \tightcitealp{rajvanshi2024saynav}; \tightcitealp{zhang2025apexnav}). Representative semantic frontier methods combine occupancy mapping, frontier scoring, and semantic fusion to decide \emph{where} evidence should guide exploration. Compared with ApexNav's target-centric fusion~\tightcitep{zhang2025apexnav}, ConsistNav studies \emph{when} accumulated evidence should become a commitment and how it should be verified, suppressed, or recovered during action.

\subsection{Classical Executive Control for ZSON}

Our work is also connected to classical planning and temporal abstraction, where persistent goals, preconditions, recovery behaviors, and action hierarchies provide explicit structure for long-horizon decision making~(\tightcitealp{ghallab2004planning}; \tightcitealp{lavalle2006planning}; \tightcitealp{sutton1999options}; \tightcitealp{kaelbling1998pomdp}). Similar ideas appear in local navigation, behavior trees, frontier exploration, and modular semantic mapping, which separate mapping, planning, and action while exposing guard conditions for when a controller should continue, recover, or switch modes~(\tightcitealp{fox1997dynamic}; \tightcitealp{colledanchise2018behavior}; \tightcitealp{yamauchi1997frontier}; \tightcitealp{chaplot2020sem}; \tightcitealp{chaplot2020ans}). We do not claim finite-state control as new; instead, our novelty lies in identifying belief-to-action inconsistency in open-vocabulary semantic navigation, defining measurable failure modes, and showing that a lightweight executive can close this gap within a modern VLM-based ObjectNav stack. Thus, ConsistNav is not a generic state machine, but an executive layer specialized for failures that emerge when semantic evidence, frontier exploration, and stop decisions interact online.

\subsection{Perception-Exploration-Control Structure for ZSON}

Recent embodied pipelines commonly organize ZSON into a modular perception--exploration--control loop, using open-vocabulary detectors, vision--language models, and promptable segmenters to obtain object evidence~(\tightcitealp{liu2024groundingdino}; \tightcitealp{li2022glip}; \tightcitealp{minderer2022owlvit}; \tightcitealp{gu2022vild}; \tightcitealp{li2023blip2}; \tightcitealp{kirillov2023sam}; \tightcitealp{zhang2023mobilesam}) while relying on frontier exploration, semantic mapping, and geometric planning for navigation. In these systems, perception proposes object evidence, mapping stores it in spatial context, and planning converts the current map into frontier or target-directed motion~(\tightcitealp{yamauchi1997frontier}; \tightcitealp{fox1997dynamic}; \tightcitealp{mur2015orbslam}; \tightcitealp{campos2021orbslam3}; \tightcitealp{mccormac2017semanticfusion}; \tightcitealp{rosinol2020kimera}). These components improve perception and search, but they do not specify how semantic evidence should be committed, recovered, or rejected during execution, especially when detections are intermittent or viewpoints are ambiguous. The execution layer still needs a policy for when to trust a weak cue, when to keep approaching it, and when to return to exploration after contradictory observations. Learning-based ObjectNav, hierarchical policies, options, and POMDP formulations can learn or model such couplings under supervision or large-scale training~(\tightcitealp{wijmans2020ddppo}; \tightcitealp{chaplot2020sem}; \tightcitealp{chaplot2020ans}; \tightcitealp{ramakrishnan2022poni}; \tightcitealp{ramrakhya2022habitatweb}; \tightcitealp{yadav2023ovrl}; \tightcitealp{ramrakhya2023pirlnav}; \tightcitealp{sutton1999options}; \tightcitealp{kaelbling1998pomdp}). ConsistNav instead targets the zero-shot setting, enforcing a lightweight reliability contract between perception and planning without retraining detector--planner-specific policies.

%% file: sections/method.tex
\section{Method}
\label{sec:method}

\subsection{Problem Formulation}
\label{sec:setting}

We study ObjectNav in previously unseen indoor scenes, where an embodied agent must find an instance of a target category from egocentric RGB-D observations and start-relative odometry. The agent outputs discrete actions under a fixed budget and is evaluated by SR and SPL. Our goal is to improve evidence-to-action consistency while keeping perception and low-level planning fixed.

\subsection{Overview}
\label{sec:model_overview}

\begin{figure}[t]
  \centering
  \includegraphics[width=\linewidth]{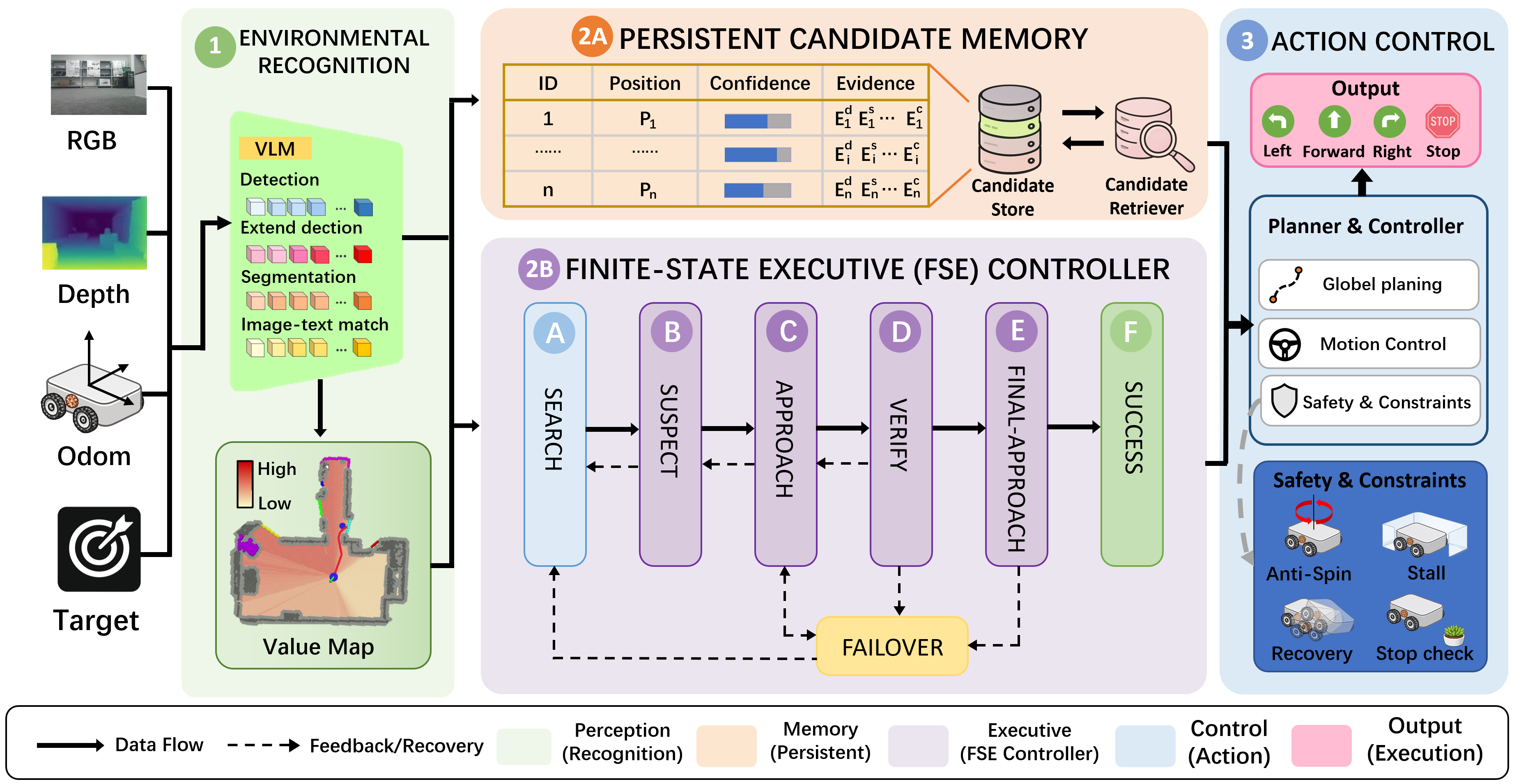}
  \caption{\textbf{ConsistNav pipeline.} \textcircled{\scriptsize\textbf{1}} Perception converts RGB-D and target cues through VLM scoring into value maps; \textcircled{\scriptsize\textbf{2A}} \textcircled{\scriptsize\textbf{2B}} planning maintains candidates and selects frontier/candidate subgoals; \textcircled{\scriptsize\textbf{3}} execution outputs \textsc{Left}, \textsc{Forward}, \textsc{Right}, and \textsc{Stop} actions through the FSE controller.}
  \label{fig:overview}
\end{figure}

ConsistNav is a perception--planning--execution framework for zero-shot ObjectNav. As shown in Figure~\ref{fig:overview}, open-vocabulary perception projects scored RGB-D evidence into semantic maps; the executive maintains persistent candidates and a finite-state controller; and the planner executes the selected frontier-conditioned or candidate-conditioned subgoal.

At step $t$, the agent receives target category $g$, RGB-D observation $o_t=(I_t,D_t)$, and pose estimate $\mathbf{x}_t=(p_t,\theta_t)$. Perception and mapping fuse scored target/non-target observations into a persistent candidate memory
\begin{equation}
  \mathcal{C}_t=\{c_1^{(t)},\dots,c_{N_t}^{(t)}\}.
\end{equation}
Each candidate stores location, confidence, semantic evidence, consistency, ITM history, and recovery flags. Conditioned on this memory, pose, progress, and failed subgoals, ConsistNav maintains a semantic control state
\begin{equation}
  q_t \in \{\stSearch,\stSuspect,\stApproach,\stVerify,\stFinalApproach,\stFailover,\stSuccess\}
\end{equation}
that gates semantic evidence before geometric planning. Given $q_t$, ConsistNav selects a frontier, approach, verification, docking, or recovery subgoal. Stability-aware action control then suppresses spin, detects stalls, bounds recovery, and authorizes \textsc{Stop} only under verified multi-cue evidence. The action output is
\begin{equation}
  a_t \in \mathcal{A}=\{\textsc{Forward},\textsc{Left},\textsc{Right},\textsc{Stop}\}.
\end{equation}
Thus, $\mathcal{C}_t$ stores accumulated evidence, $q_t$ gates planning, and $a_t$ remains in the standard ObjectNav action space.

The following subsections make these components concrete. Section~\ref{sec:fsm} details candidate memory, Section~\ref{sec:semantic_commitment} explains the Finite-State Executive Controller, and Section~\ref{sec:stability} describes stability-aware action control.

\subsection{Persistent Candidate Memory}
\label{sec:fsm}

As shown in Fig.~\ref{fig:fse_module} (left), persistent candidate memory builds a semantic candidate map and stores hypotheses by confidence, high ($0.7$--$1.0$), medium ($0.4$--$0.7$), or low ($0$--$0.4$), with last-update time. It aggregates evidence, suppression signals, and recovery status into a ranked candidate set for commitment and control.

\paragraph{Candidate representation.}
Each object hypothesis is represented as
\begin{equation}
  \scalebox{0.82}{$\displaystyle
  c_i^{(t)}=
  \Bigl(
  \mu_i^{(t)},
  \hat{c}_i^{(t)},
  n_i^{(t)},
  m_{i,+}^{(t)},
  m_{i,-}^{(t)},
  s_i^{(t)},
  \bar{\iota}_i^{(t)},
  f_i^{(t)},
  u_i^{(t)}
  \Bigr).
  $}
  \label{eq:candidate}
\end{equation}
where $\mu_i^{(t)}\in\mathbb{R}^2$ is the candidate position, $\hat{c}_i^{(t)}$ and $n_i^{(t)}$ are accumulated positive and negative evidence, $m_{i,+}^{(t)}$ and $m_{i,-}^{(t)}$ count target and non-target observations, $s_i^{(t)}$ is a consistency score, $\bar{\iota}_i^{(t)}$ stores ITM history, and $f_i^{(t)},u_i^{(t)}$ encode failure, cooldown, and later recovery status.

\paragraph{Association and belief update.}
Given a projected semantic observation $\tilde{\mu}_t$, the executive associates it with the nearest candidate and merges it when the distance is below the merge radius $r_m$.

For a matched candidate, the center is updated by exponential smoothing,
\begin{equation}
  \mu_{i^\star}^{(t+1)} = (1-\lambda)\mu_{i^\star}^{(t)} + \lambda \tilde{\mu}_t,
  \label{eq:position_update}
\end{equation}
where $\lambda\in(0,1)$ controls the contribution of the new observation; we use $\lambda=0.3$.

Confidence and negative evidence are updated asymmetrically:
\begin{equation}
  \hat{c}_i^{(t+1)} = \min\!\bigl(1,\hat{c}_i^{(t)}+\alpha\,y_t-0.12\alpha(1-y_t)\bigr),
  \qquad
  n_i^{(t+1)} = n_i^{(t)} + \beta\,z_t.
  \label{eq:belief_update}
\end{equation}
Here, $y_t$ indicates target consistency, $z_t$ indicates verify-stage failure, and $\alpha,\beta$ control confidence gain and failure penalty. This asymmetric update lets target evidence raise confidence quickly while allowing non-target or failed evidence to suppress candidates gradually.

\paragraph{Consistency score and priority.}
To decide which hypotheses can influence control, the executive first converts the memory fields into a consistency score
\begin{equation}
  s_i^{(t)}=\mathrm{clip}_{[0,1]}
  \Bigl(
  w_c \hat{c}_i^{(t)}
  + w_r \frac{m_{i,+}^{(t)}}{m_{i,+}^{(t)}+m_{i,-}^{(t)}+\varepsilon}
  + w_o \min(1,m_{i,+}^{(t)}/6)
  + w_\iota \bar{\iota}_i^{(t)}
  - w_f \phi_i^{(t)}
  \Bigr),
  \label{eq:consistency}
\end{equation}
where $\varepsilon>0$ prevents division by zero, $\phi_i^{(t)}$ summarizes failure-related penalties, and $w_c,w_r,w_o,w_\iota,w_f$ are fixed executive weights.

Candidate selection then ranks hypotheses by combining confidence, negative evidence, consistency, repeated support, and ITM confirmation, while penalizing stale, failed, or recently visited candidates.

Overall, this memory turns noisy semantic observations into a compact ranked candidate set for the executive controller.

\subsection{Finite-State Executive Controller}
\label{sec:semantic_commitment}

\begin{figure}[t]
  \centering
  \includegraphics[width=\linewidth]{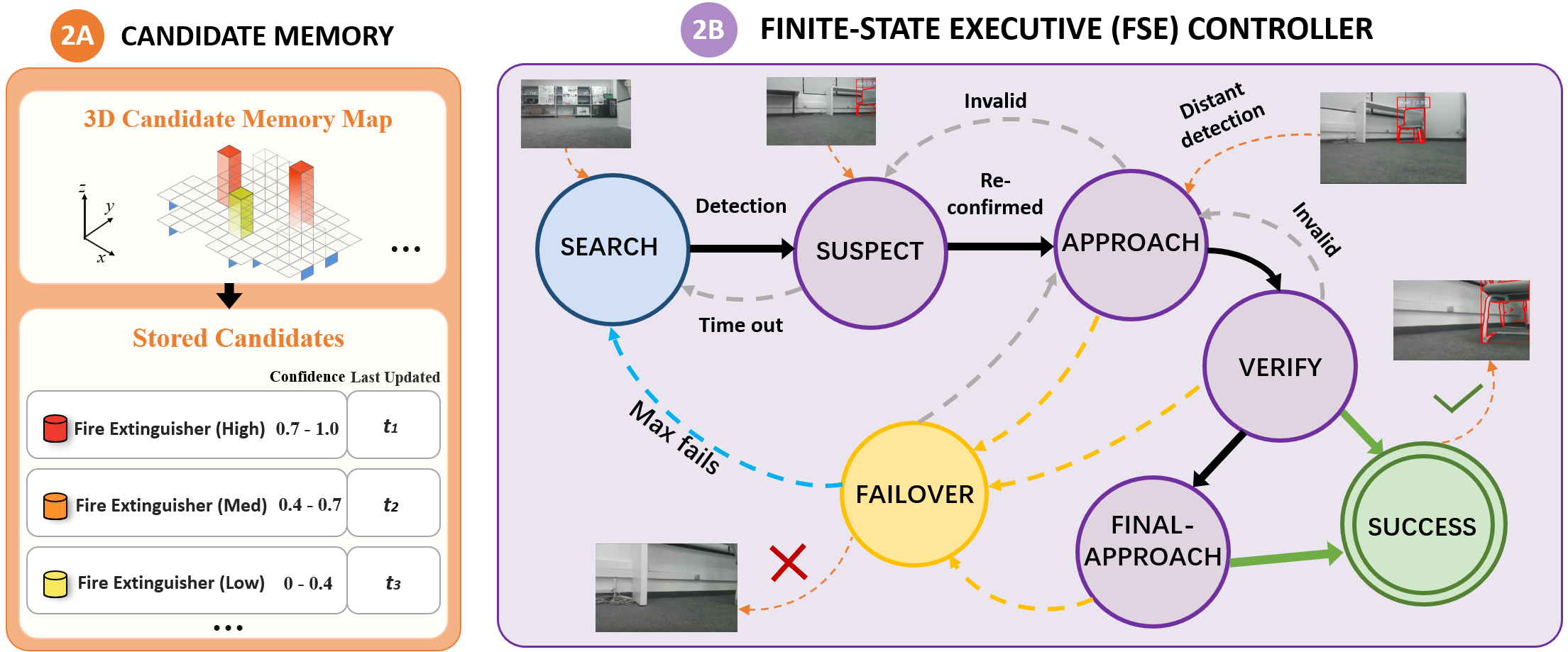}
  \caption{\textbf{Candidate Memory and FSE Controller.} Left: Candidate Memory builds/stores the semantic candidate map. Right: seven-state FSE transitions, with black/green for commitment/success, gray/yellow for invalidation/recovery, and blue for returning to search.}
  \label{fig:fse_module}
\end{figure}

As shown in Fig.~\ref{fig:fse_module} (right), the FSE Controller provides the executive flow that converts persistent candidates into committed, verified, and recoverable navigation behavior. We describe this process through the \emph{Commitment state space}, \emph{Guarded commitment chain}, and \emph{Failure loop} below.

\paragraph{Commitment state space.}
ConsistNav uses the semantic state space defined in Sec.~\ref{sec:model_overview}. These states monotonically raise commitment from frontier search to candidate holding, pursuit, close-range verification, final docking, recovery, and terminal stop, while keeping recovery as an explicit route for invalidated hypotheses.

\paragraph{Guarded commitment chain.}
The nominal chain raises commitment from open-ended search to verified termination while preserving a single active intent:
\begin{equation} \stSearch \rightarrow \stSuspect \rightarrow \stApproach \rightarrow \stVerify \rightarrow \stFinalApproach \rightarrow \stSuccess . \label{eq:fse_transition} \end{equation}
but every edge is guarded by candidate viability, rank persistence, distance, and verification evidence. A candidate is viable only if
\begin{equation} \mathcal{V}_t=\{i:f_i^{(t)}=0,\,t\ge u_i^{(t)},\,m_{i,+}^{(t)}>m_{i,-}^{(t)},\,\hat{c}_i^{(t)}\ge\tau_c,\,s_i^{(t)}\ge\tau_{\mathrm{cons}}\}, \label{eq:viable_candidates} \end{equation}
where $f_i$ marks failed hypotheses, $u_i$ is the cooldown step, $m_{i,+},m_{i,-}$ count target and non-target support, and $\hat{c}_i,s_i$ denote confidence and consistency. Viable candidates are ranked by $\pi_i^{(t)}$, which combines confidence, consistency, repeated support, ITM evidence, and stale/failure penalties. The executive enters \stSuspect{} only for a strong or persistent top candidate, advances after re-confirmation or stable rank, and switches from \stApproach{} to \stVerify{} when
\begin{equation} d(p_t,\mu_{i^\star}^{(t)}) \le r_v+\mathbb{I}[k_{\mathrm{app}}\ge N_{\mathrm{verify}}]\,r_{\mathrm{extra}}, \label{eq:approach_to_verify} \end{equation}
or when approach progress saturates. Verification succeeds only under a bounded semantic-geometric gate:
\begin{equation}
  h_t\ge2,\quad d_t^{\mathrm{best}}\le r_{\mathrm{stop}},\quad \kappa_t\ge\tau_{\mathrm{conf}},\quad
  m_{i,+}^{(t)}\ge M_{\mathrm{obs}},\quad m_{i,+}^{(t)}>m_{i,-}^{(t)},\quad \bar{\eta}_t\ge\tau_{\mathrm{itm}} \label{eq:verified_gate}
\end{equation}
so \texttt{Stop} requires agreement among target hits, geometry, confidence, observation count, and ITM evidence.

\paragraph{Failure loop.}
Complementary to semantic escalation, ConsistNav defines a recovery route for commitments that should be weakened rather than blindly continued:
\begin{equation}
  \mathcal{Q}_{\mathrm{commit}} \rightarrow \stFailover \rightarrow \mathcal{Q}_{\mathrm{return}} .
  \label{eq:failure_route}
\end{equation}
The committed and returnable state sets are
\begin{equation}
  \left\{
  \begin{aligned}
  \mathcal{Q}_{\mathrm{commit}} &= \{\stSuspect,\stApproach,\stVerify,\stFinalApproach\},\\
  \mathcal{Q}_{\mathrm{return}} &= \{\stSearch,\stApproach,\stVerify\}.
  \end{aligned}
  \right.
  \label{eq:failure_update}
\end{equation}
It then either completes docking for a recently verified close candidate, switches to the best unfrozen candidate, or returns to frontier search when no viable candidate or failover budget remains.

Overall, ConsistNav converts open-vocabulary semantic evidence into stable navigation commitment. It avoids over-trusting single-frame detections while also preventing brief occlusions or missed detections from prematurely discarding a plausible target.

\subsection{Stability-Aware Action Control}
\label{sec:stability}

Stability-Aware Action Control makes planner commands state-conditioned, so the same planner follows different executive intents. In \stSearch{}, ConsistNav follows frontier exploration; in \stApproach{} and \stFinalApproach{}, it selects a candidate-conditioned subgoal:
\begin{equation}
  \gamma_t^\star = \arg\min_{\gamma\in \mathcal{G}_t}
  \Bigl[
  d(\gamma,\mu_{i^\star}^{(t)})
  + \lambda_v\,\mathbb{I}[\gamma\in\mathcal{R}_t]
  + \lambda_f\,\mathbb{I}[\gamma\in\mathcal{F}_t]
  \Bigr],
  \label{eq:subgoal}
\end{equation}
where $d(\gamma,\mu_{i^\star}^{(t)})$ is the planner distance to the active candidate, $\mathcal{R}_t$ and $\mathcal{F}_t$ denote visited and failed regions, and $\lambda_v,\lambda_f$ weight the corresponding penalties. The objective preserves candidate commitment while discouraging repeated or failed subgoals; four guards then stabilize execution.

\paragraph{Anti-spin control.}
Let $\Delta p_t=\|p_t-p_{t-1}\|$ and $\Delta\theta_t=|\theta_t-\theta_{t-1}|$. The executive maintains a spin budget $b_t^{\mathrm{spin}}$ that increases when $\Delta\theta_t>0$ but $\Delta p_t<\delta_{\mathrm{move}}$, resets after sufficient translation, and suppresses additional pure turns once $b_t^{\mathrm{spin}}>B_{\mathrm{spin}}$ by requiring the current or resampled subgoal to satisfy a minimum translation margin.

\paragraph{Stall detection.}
The controller uses a stall counter $s_t$ for physical progress and a best-distance tracker $d_t^{\mathrm{best}}=\min_{\tau\le t} d(p_\tau,\mu_{i^\star})$ for semantic pursuit. If $\Delta p_t<\delta_{\mathrm{move}}$ or $d_t^{\mathrm{best}}$ fails to improve by margin $\epsilon_d$ for $K_s$ steps in committed states, the guard triggers verification, recovery, or failover according to candidate strength and proximity.

\paragraph{Bounded recovery.}
Recovery is finite and updates memory. Failed pursuit increases the active candidate's negative evidence $n_i^-$ and failure count $f_i$, decays its confidence/consistency, and records the region in $\mathcal{F}_t$. If no progress is observed within the $B_r$-step escape budget, \stFailover{} switches candidate, refreshes the docking point, returns to search, or falls back to frontier exploration.

\paragraph{Verified stop check.}
ConsistNav authorizes \textsc{Stop} only through a semantic-geometric gate that aggregates target hits $h_t$, best distance $d_t^{\mathrm{best}}$, maximum confidence $\kappa_t$, and average ITM score $\eta_t^{\mathrm{ITM}}$. The gate requires target evidence to dominate non-target evidence; unexpected \textsc{Stop} outputs outside \stSuccess{} are intercepted as recovery or heading reset.

Together, these guards make the action layer explicit: rotation is bounded by $b_t^{\mathrm{spin}}$, pursuit by $d_t^{\mathrm{best}}$, recovery by $(n_i^-,f_i,B_r)$, and termination by $(h_t,d_t^{\mathrm{best}},\kappa_t,\eta_t^{\mathrm{ITM}})$.

%% file: sections/experiment.tex
\section{Experiments}
\label{sec:exp}

We describe the benchmarks and implementation used for controlled evaluation, compare with prior ObjectNav methods, analyze failure modes, validate executive components through ablations, and report real-world deployment results.

\subsection{Benchmarks and Implementation Details}
\label{sec:exp_setup}

\paragraph{Datasets.}
We evaluate under the Habitat ObjectNav protocol~(\tightcitealp{savva2019habitat}; \tightcitealp{batra2020objectnav}) on three benchmark settings: MP3D~\tightcitep{chang2017matterport3d}, HM3Dv1~\tightcitep{ramakrishnan2021hm3d}, and HM3Dv2. HM3Dv1 uses the HM3D-Semantics-v0.1 validation split with 2000 episodes over 20 scenes and 6 goal categories; HM3Dv2 uses the v0.2 validation split with 1000 episodes over 36 scenes and the same categories; MP3D uses 2195 validation episodes over 11 Matterport3D scenes and 21 categories. Together, these benchmarks test transfer across scene scale, reconstruction source, and category distribution.

\paragraph{Evaluation metrics.}
We report \SR{} and \SPL{}, the two standard metrics for ObjectNav~\tightcitep{anderson2018evaluation}. \SR{} measures the fraction of episodes in which the agent reaches the target and issues a valid stop, while \SPL{} further discounts successful episodes by path inefficiency. Let $s_i$ denote success, $l_i$ the executed path length, and $l_i^\star$ the shortest feasible path length for episode $i$; higher values indicate better performance for both metrics. The metrics are defined as follows:
\begin{equation}
\SR = \frac{1}{N}\sum_{i=1}^{N}s_i,\qquad
\SPL = \frac{1}{N}\sum_{i=1}^{N}
s_i\frac{l_i^\star}{\max(l_i,l_i^\star)} .
\label{eq:exp_metrics}
\end{equation}

\paragraph{Implementation details.}
All methods follow the same Habitat protocol with a $[0,5]\,\mathrm{m}$ sensing range, $0.2\,\mathrm{m}$ success radius, and $0.88\,\mathrm{m}$ RGB-D camera height. To isolate executive control, all variants use the same perception--mapping--planning stack with YOLO-World/GroundingDINO, MobileSAM, and BLIP-2. We fix all executive thresholds across datasets, including $\tau_c=0.15$, $\tau_{\text{cons}}=0.42$, $r_v=0.8\,\mathrm{m}$, $r_{\text{stop}}=0.28\,\mathrm{m}$, $\tau_{\text{conf}}=0.30$, and $\tau_{\text{itm}}=0.12$. The internal $r_{\text{stop}}$ gate only triggers candidate-centered stop verification; success still requires the Habitat evaluator's $0.2\,\mathrm{m}$ valid-\textsc{Stop} criterion.

\subsection{Comparison with State-of-the-art}
\label{sec:main_results}

Table~\ref{tab:main} compares ConsistNav with representative ObjectNav methods on HM3Dv2, HM3Dv1, and MP3D. Prior rows are literature-reported contextual baselines; the primary evidence is a deterministic same-stack comparison against the Non-executive baseline, which keeps the detector, mapper, planner, simulator protocol, episode list, and action budget fixed. Under this setting, ConsistNav improves \SR{}/\SPL{} by $+8.0\%/+3.2\%$ on HM3Dv2, $+3.6\%/+1.8\%$ on HM3Dv1, and $+11.4\%/+7.9\%$ on MP3D. Among the methods listed in Table~\ref{tab:main}, it obtains the strongest reported results on all three benchmarks, indicating that persistent memory, finite-state commitment, verified stopping, and bounded failover improve reliability without changing the underlying stack.

\begin{table}[t]
\centering
\setlength{\abovecaptionskip}{4pt}
\setlength{\belowcaptionskip}{6pt}
\caption{\small ObjectNav results on HM3Dv2, HM3Dv1, and MP3D. Prior rows are literature-reported contextual baselines; controlled same-stack comparisons are described in the text and ablations.}
\label{tab:main}
{\small
\setlength{\tabcolsep}{0pt}
\renewcommand{\arraystretch}{1.42}
\begin{tabular*}{\linewidth}{@{}
>{\raggedright\arraybackslash}p{0.345\linewidth}
>{\centering\arraybackslash}p{0.13\linewidth}
>{\centering\arraybackslash}p{0.10\linewidth}
*{6}{>{\centering\arraybackslash}p{0.066\linewidth}}
@{}}
\toprule
& & & \multicolumn{2}{c}{HM3Dv2} & \multicolumn{2}{c}{HM3Dv1} & \multicolumn{2}{c}{MP3D} \\
\cmidrule(lr){4-5} \cmidrule(lr){6-7} \cmidrule(lr){8-9}
Method & Unsupervised & Zero-shot & \SR{} & \SPL{} & \SR{} & \SPL{} & \SR{} & \SPL{} \\
\midrule
PONI~(\tightcitealp{ramakrishnan2022poni}) & No & No & -- & -- & -- & -- & 31.8 & 12.1 \\
ProcTHOR~(\tightcitealp{deitke2022procthor}) & No & No & -- & -- & 54.4 & 31.8 & -- & -- \\
\midrule
ZSON~(\tightcitealp{majumdar2022zson}) & Yes & No & -- & -- & 25.5 & 12.6 & 15.3 & 4.8 \\
ProcTHOR-ZS~(\tightcitealp{deitke2022procthor}) & Yes & No & -- & -- & 13.2 & 7.7 & -- & -- \\
\midrule
OpenFMNav~(\tightcitealp{kuang2024openfmnav}) & Yes & Yes & -- & -- & 54.9 & 24.4 & 37.2 & 15.7 \\
InstructNav~(\tightcitealp{long2024instructnav}) & Yes & Yes & 58.0 & 20.9 & -- & -- & -- & -- \\
VLFM~(\tightcitealp{yokoyama2024vlfm}) & Yes & Yes & 63.6 & 32.5 & 52.5 & 30.4 & 36.4 & 17.5 \\
TriHelper~(\tightcitealp{zhang2024trihelper}) & Yes & Yes & -- & -- & 56.5 & 25.3 & -- & -- \\
SG-Nav~(\tightcitealp{yin2024sgnav}) & Yes & Yes & 49.6 & 25.5 & 54.0 & 24.9 & \underline{40.2} & 16.0 \\
ApexNav~(\tightcitealp{zhang2025apexnav}) & Yes & Yes & \underline{76.2} & \underline{38.0} & \underline{59.6} & \underline{33.0} & 39.2 & \underline{17.8} \\
\midrule
\textbf{ConsistNav (ours)} & Yes & Yes & \textbf{84.2} & \textbf{41.2} & \textbf{63.2} & \textbf{34.8} & \textbf{50.6} & \textbf{25.7} \\
\bottomrule
\end{tabular*}}
\end{table}

\begin{figure}[t]
\centering
\includegraphics[width=\linewidth]{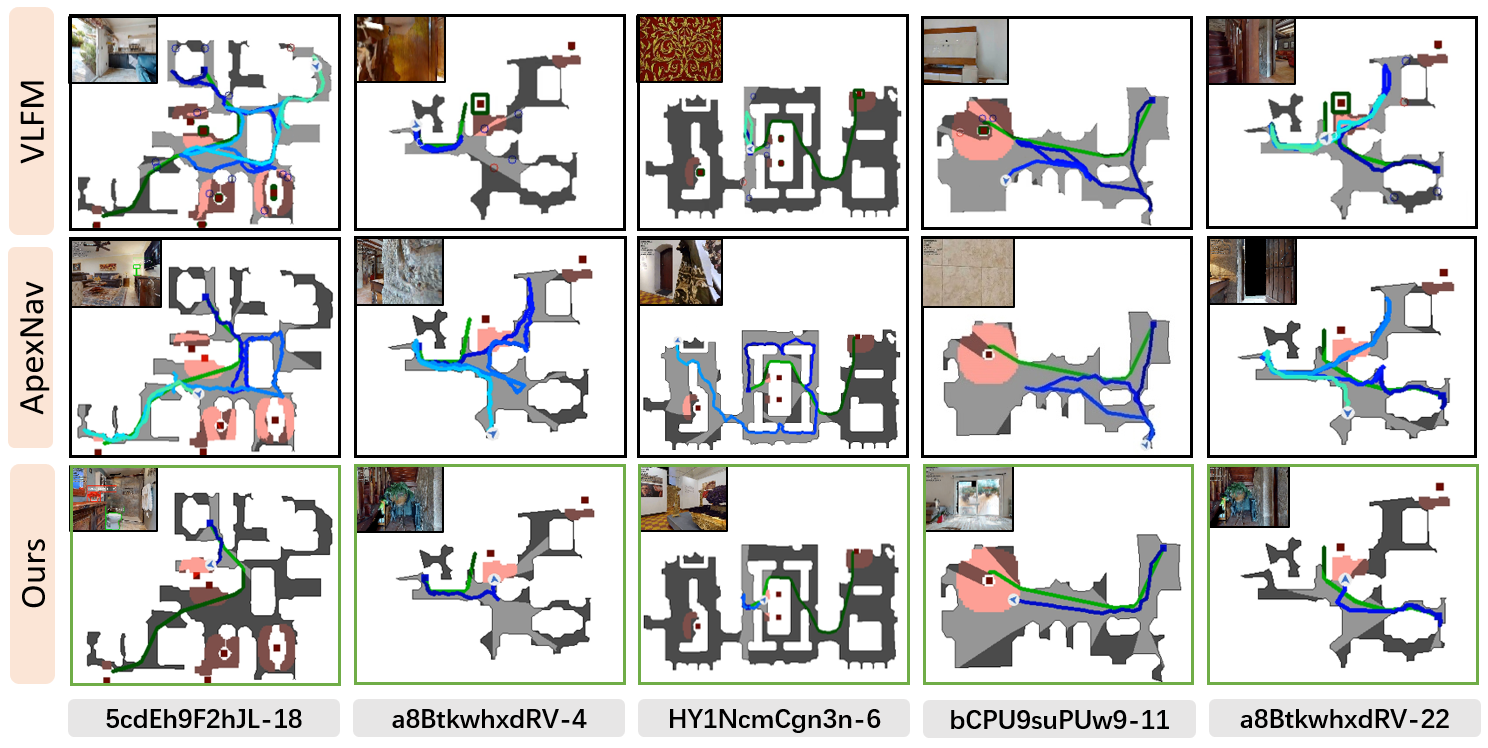}
\caption{\textbf{Simulation results on HM3Dv2.} Qualitative comparison of ConsistNav, VLFM, and ApexNav. Each column shows one episode; green/blue paths denote reference/agent trajectories, and green/black frames denote success/failure.}
\label{fig:exp-sim}
\end{figure}

\subsection{Failure Cause Analysis}
\label{sec:failure_analysis}

\begin{figure}[t]
\centering
\includegraphics[width=\linewidth]{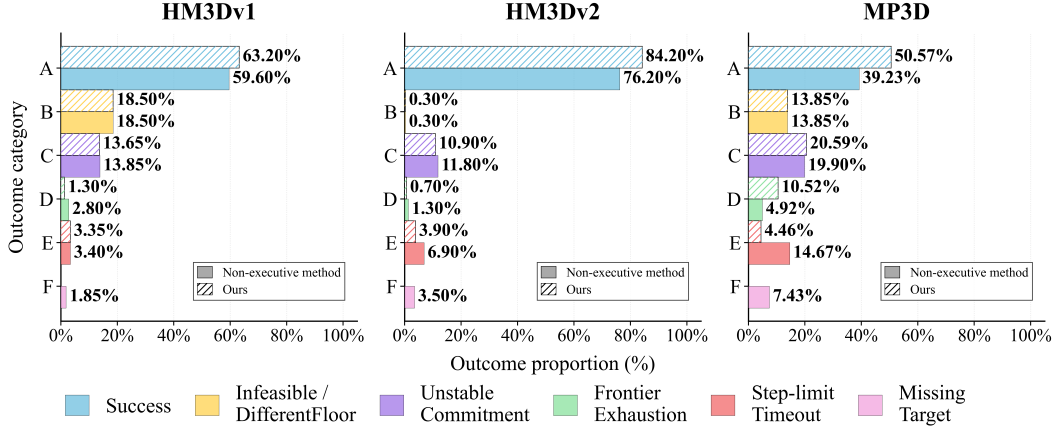}
\caption{\textbf{Failure-cause comparison.} Outcome statistics for the Non-executive method and \ours{} on HM3Dv1, HM3Dv2, and MP3D, covering verified success and five residual failure modes.}
\label{fig:failure_cause}
\end{figure}

We assign each episode to one outcome after inspecting the trajectory, semantic detections, stop decision, and target passage. As shown in Fig.~\ref{fig:failure_cause}, we use six categories: (1) \emph{Success}, reaching and verifying the target; (2) \emph{Infeasible / Different Floor}, unreachable or topologically separated targets; (3) \emph{Unstable Commitment}, observed evidence followed by abandonment or unreliable stopping; (4) \emph{Frontier Exhaustion}, no remaining frontier goals; (5) \emph{Step-limit Timeout}, search or recovery exceeding the action budget; and (6) \emph{Missing Target}, passing the target without verification.

Figure~\ref{fig:failure_cause} compares ConsistNav with the Non-executive method and shows that the executive shifts controllable failures toward verified success. On HM3Dv2 and MP3D, success rises by $+8.00\%$ and $+11.34\%$ while timeout drops by $3.00\%$ and $10.21\%$, indicating that memory, guarded commitment, and verified stopping sustain promising pursuits and reduce missed or premature termination. MP3D frontier exhaustion increases from $4.92\%$ to $10.52\%$, reflecting a conservative trade-off: weak candidates become explicit search failures rather than unstable commitments, while infeasible and late-discovery cases remain dataset-level limits.

\subsection{Ablation Study}
\label{sec:ablation}

\begin{table}[t]
\centering
\begin{minipage}[t]{0.58\linewidth}
\centering
\caption{Executive ablation on HM3Dv2 under the same underlying stack. PCM, FSEC, and SAAC denote persistent candidate memory, Finite-State Executive Controller, and stability-aware action control.}
\label{tab:ablation}
\scriptsize
\setlength{\tabcolsep}{2.5pt}
\renewcommand{\arraystretch}{0.94}
\resizebox{\linewidth}{!}{%
\begin{tabular}{lccccc}
\toprule
Variant & PCM & FSEC & SAAC & \SR{} & \SPL{} \\
\midrule
Non-executive baseline & \ding{55} & \ding{55} & \ding{55} & 76.20 & 38.00 \\
+ Persistent Candidate Memory & \ding{51} & \ding{55} & \ding{55} & 77.10 & 39.60 \\
+ Finite-State Executive Controller & \ding{51} & \ding{51} & \ding{55} & 81.00 & 40.75 \\
+ Stability-Aware Action Control & \ding{51} & \ding{51} & \ding{51} & \textbf{84.20} & \textbf{41.24} \\
\bottomrule
\end{tabular}}
\end{minipage}
\hfill
\begin{minipage}[t]{0.38\linewidth}
\centering
\caption{Real-world deployment results with 10 physical trials per target.}
\label{tab:real_quant}
\scriptsize
\setlength{\tabcolsep}{2pt}
\renewcommand{\arraystretch}{0.84}
\resizebox{\linewidth}{!}{%
\begin{tabular}{lccc}
\toprule
Target & Episodes & Success & Time(avg) \\
\midrule
 Chair & 10 & 100 & 157s \\
 Plant & 10 & 100 & 140s \\
 Couch & 10 & 100 & 164s \\
 TV & 10 & 100 & 160s \\
\midrule
 Overall & 40 & 100 & 155s \\
\bottomrule
\end{tabular}}
\end{minipage}
\end{table}

\paragraph{Ablation analysis.}
Table~\ref{tab:ablation} shows progressive improvement as the three executive components are added, indicating that ConsistNav benefits from the interaction between memory, executive control, and action stabilization rather than a single heuristic. The monotonic trend also suggests that each module removes a different bottleneck: noisy semantic evidence, unstable commitment, and low-level action inconsistency. This staged comparison is therefore more diagnostic than a single full-system comparison, because it separates whether gains come from remembering candidates, deciding when to pursue them, or executing the pursuit robustly. \textbf{(a) Persistent Candidate Memory.} PCM provides the first gain of $+0.90\%$ \SR{} and $+1.60\%$ \SPL{} by stabilizing cross-frame evidence and suppressing transient detections, although memory alone cannot decide when to commit, recover, or stop. \textbf{(b) Finite-State Executive Controller.} FSEC contributes the largest improvement, adding $+3.90\%$ \SR{} and $+1.15\%$ \SPL{}, showing that guarded state transitions are the main mechanism for turning stored candidates into reliable commitments. \textbf{(c) Stability-Aware Action Control.} SAAC further improves \SR{} by $+3.20\%$ and \SPL{} by $+0.49\%$ by handling spin, stall, verified stopping, and failover after target selection. Overall, the full executive achieves $+8.00\%$ \SR{} and $+3.24\%$ \SPL{} over the Non-executive baseline, confirming that the three modules address complementary stages of the evidence-to-action pipeline.

\subsection{Real-world Deployment}
\label{sec:real_world}

\begin{figure}[t]
\centering
\includegraphics[width=\linewidth]{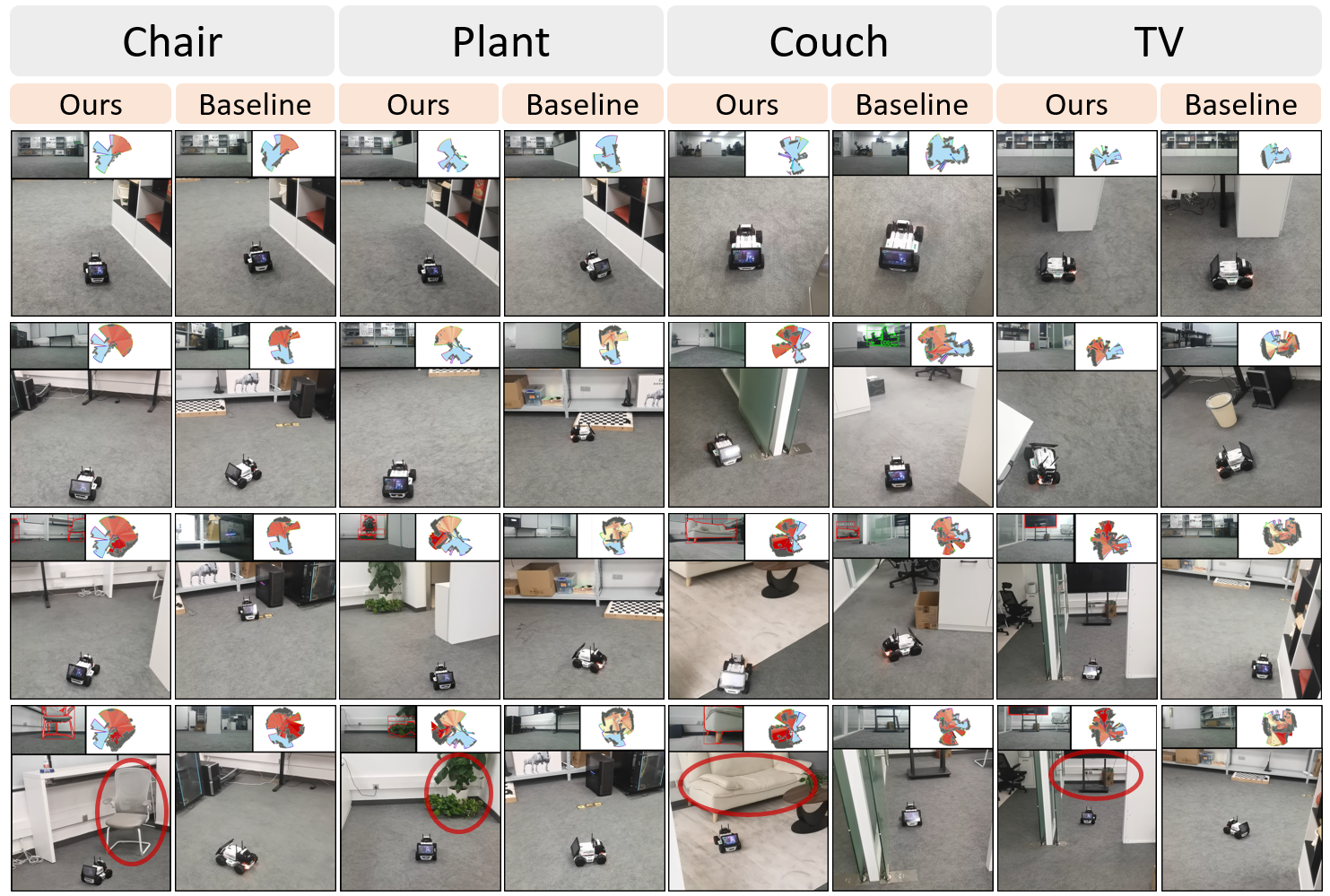}
\caption{\textbf{Real-world deployment comparison.} Visual comparison of the Non-executive baseline and ConsistNav on four target tasks using the AgileX LIMO platform. The results illustrate that ConsistNav maintains target hypotheses, verifies close-range evidence, and stops reliably under real sensor and timing conditions.}
\label{fig:exp1}
\end{figure}

We deploy ConsistNav on an AgileX LIMO robot and run inference on an NVIDIA RTX 4090 workstation. Figure~\ref{fig:exp1} provides qualitative real-world comparisons with a Non-executive baseline, showing representative navigation behaviors under real sensing and control latency. Table~\ref{tab:real_quant} reports the physical-scene evaluation: 40 tasks are conducted in a $10{\times}10\,\mathrm{m}^2$ office scene and a $30{\times}10\,\mathrm{m}^2$ constructed home scene across chair, plant, couch, and TV targets, validating the robustness of the deployed system.

%% file: sections/conclusion.tex
\section{Conclusion}
\label{sec:conclusion}

We presented ConsistNav, a training-free semantic executive that achieves state-of-the-art zero-shot ObjectNav performance without relying on online large language models. ConsistNav closes the action consistency gap by combining persistent candidate memory, the Finite-State Executive Controller, verified stopping, and bounded failover, while keeping the perception, mapping, frontier planning, and local control stack fixed. Across HM3D and MP3D, it improves both goal acquisition and path efficiency, and ablations show that each executive component contributes complementary gains. These results highlight explicit executive structure as a decisive ingredient for robust open-vocabulary navigation and a practical path toward deployable embodied agents.

\paragraph{Limitations.}
Real-world inference currently runs on a remote PC rather than fully onboard the robot. The results also depend on coordination between the simulator, planner, and verified-stop interface, reflecting the sensitivity of ObjectNav benchmarks to execution-level design.

\paragraph{Future work.}
Future work will study adaptive transition thresholds while preserving the explicit executive structure, and extend ConsistNav to multi-target or instruction-conditioned navigation where commitment must reason over competing semantic goals.

%% file: references.bib
@article{batra2020objectnav,
  title     = {{ObjectNav} Revisited: On Evaluation of Embodied Agents
               Navigating to Objects},
  author    = {Batra, Dhruv and Gokaslan, Aaron and Kembhavi, Aniruddha and
               Maksymets, Oleksandr and Mottaghi, Roozbeh and Savva, Manolis
               and Toshev, Alexander and Wijmans, Erik},
  journal   = {arXiv preprint arXiv:2006.13171},
  year      = {2020}
}

@inproceedings{savva2019habitat,
  title     = {{Habitat}: {A} Platform for Embodied {AI} Research},
  author    = {Savva, Manolis and Kadian, Abhishek and Maksymets, Oleksandr and
               Zhao, Yili and Wijmans, Erik and Jain, Bhavana and Straub, Julian
               and Liu, Jia and Koltun, Vladlen and Malik, Jitendra and
               Parikh, Devi and Batra, Dhruv},
  booktitle = {Proceedings of the IEEE/CVF International Conference on
               Computer Vision (ICCV)},
  year      = {2019}
}

@inproceedings{ramakrishnan2021hm3d,
  title     = {{Habitat-Matterport 3D Dataset (HM3D)}: 1000 Large-scale
               3D Environments for Embodied {AI}},
  author    = {Ramakrishnan, Santhosh Kumar and Gokaslan, Aaron and Wijmans, Erik
               and Maksymets, Oleksandr and Clegg, Alexander and Turner, John M.
               and Undersander, Eric and Galuba, Wojciech and Westbury, Andrew
               and Chang, Angel X. and Savva, Manolis and Zhao, Yili and
               Batra, Dhruv},
  booktitle = {Proceedings of the Neural Information Processing Systems
               Track on Datasets and Benchmarks (NeurIPS Datasets)},
  year      = {2021}
}

@inproceedings{chang2017matterport3d,
  title     = {Matterport3D: Learning from {RGB-D} Data in Indoor Environments},
  author    = {Chang, Angel X. and Dai, Angela and Funkhouser, Thomas and
               Halber, Maciej and Niessner, Matthias and Savva, Manolis and
               Song, Shuran and Zeng, Andy and Zhang, Yinda},
  booktitle = {Proceedings of the International Conference on 3D Vision (3DV)},
  year      = {2017}
}

@inproceedings{wijmans2020ddppo,
  title     = {{DD-PPO}: Learning Near-Perfect {PointGoal} Navigators from
               2.5 Billion Frames},
  author    = {Wijmans, Erik and Kadian, Abhishek and Morcos, Ari and Lee,
               Stefan and Essa, Irfan and Parikh, Devi and Savva, Manolis
               and Batra, Dhruv},
  booktitle = {International Conference on Learning Representations (ICLR)},
  year      = {2020}
}

@inproceedings{chaplot2020sem,
  title     = {Object Goal Navigation using Goal-Oriented Semantic Exploration},
  author    = {Chaplot, Devendra Singh and Gandhi, Dhiraj and Gupta, Abhinav
               and Salakhutdinov, Ruslan},
  booktitle = {Advances in Neural Information Processing Systems (NeurIPS)},
  year      = {2020}
}

@inproceedings{ramakrishnan2022poni,
  title     = {{PONI}: Potential Functions for {ObjectGoal} Navigation with
               Interaction-free Learning},
  author    = {Ramakrishnan, Santhosh K. and Chaplot, Devendra Singh and
               Al-Halah, Ziad and Malik, Jitendra and Grauman, Kristen},
  booktitle = {Proceedings of the IEEE/CVF Conference on Computer Vision and
               Pattern Recognition (CVPR)},
  year      = {2022}
}

@inproceedings{ramrakhya2022habitatweb,
  title     = {Habitat-Web: Learning Embodied Object-Search Strategies from
               Human Demonstrations at Scale},
  author    = {Ramrakhya, Ram and Undersander, Eric and Batra, Dhruv and
               Das, Abhishek},
  booktitle = {Proceedings of the IEEE/CVF Conference on Computer Vision and
               Pattern Recognition (CVPR)},
  year      = {2022}
}

@article{yadav2023ovrl,
  title     = {{OVRL-V2}: A simple state-of-art baseline for {ImageNav} and
               {ObjectNav}},
  author    = {Yadav, Karmesh and Ramrakhya, Ram and Majumdar, Arjun and
               Yokoyama, Naoki and Baevski, Alexei and Kira, Zsolt and
               Maksymets, Oleksandr and Batra, Dhruv},
  journal   = {arXiv preprint arXiv:2303.07798},
  year      = {2023}
}

@inproceedings{khandelwal2022clip,
  title     = {Simple but Effective: {CLIP} Embeddings for Embodied {AI}},
  author    = {Khandelwal, Apoorv and Weihs, Luca and Mottaghi, Roozbeh and
               Kembhavi, Aniruddha},
  booktitle = {Proceedings of the IEEE/CVF Conference on Computer Vision and
               Pattern Recognition (CVPR)},
  year      = {2022}
}

@inproceedings{majumdar2022zson,
  title     = {{ZSON}: Zero-Shot Object-Goal Navigation using Multimodal Goal
               Embeddings},
  author    = {Majumdar, Arjun and Aggarwal, Gunjan and Devnani, Bhavika and
               Hoffman, Judy and Batra, Dhruv},
  booktitle = {Advances in Neural Information Processing Systems (NeurIPS)},
  year      = {2022}
}

@inproceedings{gadre2023cows,
  title     = {{COWS} on {PASTURE}: Baselines and Benchmarks for
               Language-Driven Zero-Shot Object Navigation},
  author    = {Gadre, Samir Yitzhak and Wortsman, Mitchell and Ilharco,
               Gabriel and Schmidt, Ludwig and Song, Shuran},
  booktitle = {Proceedings of the IEEE/CVF Conference on Computer Vision and
               Pattern Recognition (CVPR)},
  year      = {2023}
}

@inproceedings{yokoyama2024vlfm,
  title     = {{VLFM}: Vision-Language Frontier Maps for Zero-Shot Semantic
               Navigation},
  author    = {Yokoyama, Naoki and Ha, Sehoon and Batra, Dhruv and Wang,
               Jiuguang and Bucher, Bernadette},
  booktitle = {Proceedings of the IEEE International Conference on Robotics
               and Automation (ICRA)},
  year      = {2024}
}

@inproceedings{yu2023l3mvn,
  title     = {{L3MVN}: Leveraging Large Language Models for Visual Target
               Navigation},
  author    = {Yu, Bangguo and Kasaei, Hamidreza and Cao, Ming},
  booktitle = {Proceedings of the IEEE/RSJ International Conference on
               Intelligent Robots and Systems (IROS)},
  year      = {2023}
}

@inproceedings{shah2022lmnav,
  title     = {{LM-Nav}: Robotic Navigation with Large Pretrained Models of
               Language, Vision, and Action},
  author    = {Shah, Dhruv and Osi{\'n}ski, B{\l}a{\.z}ej and Ichter, Brian
               and Levine, Sergey},
  booktitle = {Proceedings of the Conference on Robot Learning (CoRL)},
  year      = {2022}
}

@article{zhang2025apexnav,
  title         = {{ApexNav}: An Adaptive Exploration Strategy for Zero-Shot
                   Object Navigation with Target-centric Semantic Fusion},
  author        = {Zhang, Mingjie and Du, Yuheng and Wu, Chengkai and Zhou,
                   Jinni and Qi, Zhenchao and Ma, Jun and Zhou, Boyu},
  journal       = {IEEE Robotics and Automation Letters},
  year          = {2025},
  eprint        = {2504.14478},
  archivePrefix = {arXiv}
}

@inproceedings{yamauchi1997frontier,
  title     = {A Frontier-Based Approach for Autonomous Exploration},
  author    = {Yamauchi, Brian},
  booktitle = {Proceedings of the IEEE International Symposium on
               Computational Intelligence in Robotics and Automation (CIRA)},
  year      = {1997}
}

@inproceedings{li2023blip2,
  title     = {{BLIP-2}: Bootstrapping Language-Image Pre-Training with
               Frozen Image Encoders and Large Language Models},
  author    = {Li, Junnan and Li, Dongxu and Savarese, Silvio and Hoi,
               Steven},
  booktitle = {Proceedings of the International Conference on Machine
               Learning (ICML)},
  year      = {2023}
}

@inproceedings{liu2024groundingdino,
  title     = {Grounding {DINO}: Marrying {DINO} with Grounded Pre-Training
               for Open-Set Object Detection},
  author    = {Liu, Shilong and Zeng, Zhaoyang and Ren, Tianhe and Li,
               Feng and Zhang, Hao and Yang, Jie and Jiang, Qing and Li,
               Chunyuan and Yang, Jianwei and Su, Hang and Zhu, Jun and
               Zhang, Lei},
  booktitle = {Proceedings of the European Conference on Computer Vision
               (ECCV)},
  year      = {2024}
}

@article{zhang2023mobilesam,
  title     = {Faster Segment Anything: Towards Lightweight {SAM} for Mobile
               Applications},
  author    = {Zhang, Chaoning and Han, Dongshen and Qiao, Yu and Kim, Jung
               Uk and Bae, Sung-Ho and Lee, Seungkyu and Hong, Choong Seon},
  journal   = {arXiv preprint arXiv:2306.14289},
  year      = {2023}
}

@book{ghallab2004planning,
  title     = {Automated Planning: Theory and Practice},
  author    = {Ghallab, Malik and Nau, Dana and Traverso, Paolo},
  publisher = {Elsevier},
  year      = {2004}
}

@article{sutton1999options,
  title   = {Between {MDPs} and Semi-{MDPs}: A Framework for Temporal
             Abstraction in Reinforcement Learning},
  author  = {Sutton, Richard S. and Precup, Doina and Singh, Satinder},
  journal = {Artificial Intelligence},
  volume  = {112},
  number  = {1--2},
  pages   = {181--211},
  year    = {1999}
}

@article{kaelbling1998pomdp,
  title   = {Planning and Acting in Partially Observable Stochastic Domains},
  author  = {Kaelbling, Leslie Pack and Littman, Michael L. and Cassandra,
             Anthony R.},
  journal = {Artificial Intelligence},
  volume  = {101},
  number  = {1--2},
  pages   = {99--134},
  year    = {1998}
}

@inproceedings{radford2021clip,
  title     = {Learning Transferable Visual Models from Natural Language
               Supervision},
  author    = {Radford, Alec and Kim, Jong Wook and Hallacy, Chris and
               Ramesh, Aditya and Goh, Gabriel and Agarwal, Sandhini and
               Sastry, Girish and Askell, Amanda and Mishkin, Pamela and
               Clark, Jack and Krueger, Gretchen and Sutskever, Ilya},
  booktitle = {Proceedings of the International Conference on Machine
               Learning (ICML)},
  year      = {2021}
}

@inproceedings{kirillov2023sam,
  title     = {Segment Anything},
  author    = {Kirillov, Alexander and Mintun, Eric and Ravi, Nikhila and
               Mao, Hanzi and Rolland, Chloe and Gustafson, Laura and
               Xiao, Tete and Whitehead, Spencer and Berg, Alexander C. and
               Lo, Wan-Yen and Doll{\'a}r, Piotr and Girshick, Ross},
  booktitle = {Proceedings of the IEEE/CVF International Conference on
               Computer Vision (ICCV)},
  year      = {2023}
}

@inproceedings{gu2022vild,
  title     = {Open-Vocabulary Object Detection via Vision and Language
               Knowledge Distillation},
  author    = {Gu, Xiuye and Lin, Tsung-Yi and Kuo, Weicheng and Cui, Yin},
  booktitle = {International Conference on Learning Representations (ICLR)},
  year      = {2022}
}

@inproceedings{minderer2022owlvit,
  title     = {Simple Open-Vocabulary Object Detection with Vision
               Transformers},
  author    = {Minderer, Matthias and Gritsenko, Alexey and Stone, Austin
               and Neumann, Maxim and Weissenborn, Dirk and Dosovitskiy,
               Alexey and Mahendran, Aravindh and Arnab, Anurag and
               Dehghani, Mostafa and Shen, Zhuoran and Wang, Xiao and
               Zhai, Xiaohua and Kipf, Thomas and Houlsby, Neil},
  booktitle = {Proceedings of the European Conference on Computer Vision
               (ECCV)},
  year      = {2022}
}

@inproceedings{li2022glip,
  title     = {Grounded Language-Image Pre-Training},
  author    = {Li, Liunian Harold and Zhang, Pengchuan and Zhang, Haotian
               and Yang, Jianwei and Li, Chunyuan and Zhong, Yiwu and
               Wang, Lijuan and Yuan, Lu and Zhang, Lei and Hwang,
               Jenq-Neng and Chang, Kai-Wei and Gao, Jianfeng},
  booktitle = {Proceedings of the IEEE/CVF Conference on Computer Vision
               and Pattern Recognition (CVPR)},
  year      = {2022}
}

@inproceedings{zhou2023esc,
  title     = {{ESC}: Exploration with Soft Commonsense Constraints for
               Zero-shot Object Navigation},
  author    = {Zhou, Kaiwen and Zheng, Kaizhi and Pryor, Connor and Shen,
               Yilin and Jin, Hongxia and Getoor, Lise and Wang, Xin Eric},
  booktitle = {Proceedings of the International Conference on Machine
               Learning (ICML)},
  year      = {2023}
}

@inproceedings{rajvanshi2024saynav,
  title     = {{SayNav}: Grounding Large Language Models for Dynamic
               Planning to Navigation in New Environments},
  author    = {Rajvanshi, Abhinav and Sikka, Karan and Lin, Xiao and Lee,
               Bhoram and Chiu, Han-Pang and Velasquez, Alvaro},
  booktitle = {Proceedings of the International Conference on Automated
               Planning and Scheduling (ICAPS)},
  year      = {2024}
}

@inproceedings{kuang2024openfmnav,
  title     = {{OpenFMNav}: Towards Open-Set Zero-Shot Object Navigation via
               Vision-Language Foundation Models},
  author    = {Kuang, Yuxuan and Lin, Hai and Jiang, Meng},
  booktitle = {Findings of the Association for Computational Linguistics:
               NAACL 2024},
  pages     = {338--351},
  year      = {2024}
}

@article{long2024instructnav,
  title   = {{InstructNav}: Zero-shot System for Generic Instruction
             Navigation in Unexplored Environment},
  author  = {Long, Yuxing and Cai, Wenzhe and Wang, Hongcheng and Zhan,
             Guanqi and Dong, Hao},
  journal = {arXiv preprint arXiv:2406.04882},
  year    = {2024}
}

@inproceedings{zhang2024trihelper,
  title     = {{TriHelper}: Zero-Shot Object Navigation with Dynamic
               Assistance},
  author    = {Zhang, Lingfeng and Zhang, Qiang and Wang, Hao and Xiao,
               Erjia and Jiang, Zixuan and Chen, Honglei and Xu, Renjing},
  booktitle = {Proceedings of the IEEE/RSJ International Conference on
               Intelligent Robots and Systems (IROS)},
  year      = {2024}
}

@inproceedings{yin2024sgnav,
  title   = {{SG-Nav}: Online 3D Scene Graph Prompting for LLM-based
             Zero-shot Object Navigation},
  author  = {Yin, Hang and Xu, Xiuwei and Wu, Zhenyu and Zhou, Jie and Lu,
             Jiwen},
  booktitle = {Advances in Neural Information Processing Systems (NeurIPS)},
  year    = {2024}
}

@inproceedings{chaplot2020ans,
  title     = {Learning to Explore Using Active Neural {SLAM}},
  author    = {Chaplot, Devendra Singh and Gandhi, Dhiraj and Gupta,
               Saurabh and Gupta, Abhinav and Salakhutdinov, Ruslan},
  booktitle = {International Conference on Learning Representations (ICLR)},
  year      = {2020}
}

@inproceedings{ramrakhya2023pirlnav,
  title     = {{PIRLNav}: Pretraining with Imitation and {RL} Finetuning
               for {ObjectNav}},
  author    = {Ramrakhya, Ram and Batra, Dhruv and Wijmans, Erik and Das,
               Abhishek},
  booktitle = {Proceedings of the IEEE/CVF Conference on Computer Vision
               and Pattern Recognition (CVPR)},
  year      = {2023}
}

@inproceedings{deitke2022procthor,
  title     = {{ProcTHOR}: Large-Scale Embodied {AI} Using Procedural
               Generation},
  author    = {Deitke, Matt and VanderBilt, Eli and Herrasti, Alvaro and
               Weihs, Luca and Ehsani, Kiana and Salvador, Jordi and Han,
               Winson and Kolve, Eric and Kembhavi, Aniruddha and Mottaghi,
               Roozbeh},
  booktitle = {Advances in Neural Information Processing Systems (NeurIPS)},
  year      = {2022}
}

@article{anderson2018evaluation,
  title     = {On Evaluation of Embodied Navigation Agents},
  author    = {Anderson, Peter and Chang, Angel and Chaplot, Devendra Singh
               and Dosovitskiy, Alexey and Gupta, Saurabh and Koltun,
               Vladlen and Kosecka, Jana and Malik, Jitendra and Mottaghi,
               Roozbeh and Savva, Manolis and Zamir, Amir R.},
  journal   = {arXiv preprint arXiv:1807.06757},
  year      = {2018}
}

@inproceedings{rosinol2020kimera,
  title     = {{Kimera}: An Open-Source Library for Real-Time Metric-Semantic
               Localization and Mapping},
  author    = {Rosinol, Antoni and Abate, Marcus and Chang, Yun and
               Carlone, Luca},
  booktitle = {Proceedings of the IEEE International Conference on Robotics
               and Automation (ICRA)},
  year      = {2020}
}

@inproceedings{mccormac2017semanticfusion,
  title     = {{SemanticFusion}: Dense 3{D} Semantic Mapping with
               Convolutional Neural Networks},
  author    = {McCormac, John and Handa, Ankur and Davison, Andrew J. and
               Leutenegger, Stefan},
  booktitle = {Proceedings of the IEEE International Conference on Robotics
               and Automation (ICRA)},
  year      = {2017}
}

@book{lavalle2006planning,
  title     = {Planning Algorithms},
  author    = {LaValle, Steven M.},
  publisher = {Cambridge University Press},
  year      = {2006}
}

@book{colledanchise2018behavior,
  title     = {Behavior Trees in Robotics and {AI}: An Introduction},
  author    = {Colledanchise, Michele and {\"O}gren, Petter},
  publisher = {CRC Press},
  year      = {2018}
}

@article{fox1997dynamic,
  title     = {The Dynamic Window Approach to Collision Avoidance},
  author    = {Fox, Dieter and Burgard, Wolfram and Thrun, Sebastian},
  journal   = {IEEE Robotics \& Automation Magazine},
  volume    = {4},
  number    = {1},
  pages     = {23--33},
  year      = {1997}
}

@article{mur2015orbslam,
  title     = {{ORB-SLAM}: A Versatile and Accurate Monocular {SLAM} System},
  author    = {Mur-Artal, Raul and Montiel, J. M. M. and Tard{\'o}s, Juan D.},
  journal   = {IEEE Transactions on Robotics},
  volume    = {31},
  number    = {5},
  pages     = {1147--1163},
  year      = {2015}
}

@article{campos2021orbslam3,
  title     = {{ORB-SLAM3}: An Accurate Open-Source Library for Visual, Visual-Inertial, and Multimap {SLAM}},
  author    = {Campos, Carlos and Elvira, Richard and Rodr{\'i}guez, Juan J. G{\'o}mez and Montiel, Jos{\'e} M. M. and Tard{\'o}s, Juan D.},
  journal   = {IEEE Transactions on Robotics},
  volume    = {37},
  number    = {6},
  pages     = {1874--1890},
  year      = {2021}
}
